\def\year{2019}\relax
\documentclass[letterpaper]{article} 
\usepackage{aaai19}  
\usepackage{times}  
\usepackage{helvet}  
\usepackage{courier}  
\usepackage{url}  
\usepackage{graphicx}  
\usepackage{amssymb}
\usepackage{mathtools}
\usepackage[]{algorithm2e}
\usepackage{listings}
\usepackage{multirow}

\begin{document}
%
\title{Fine Grained Classification of Personal Data Entities}
\author{Riddhiman Dasgupta,
Balaji Ganesan,
Aswin Kannan,
Berthold Reinwald,
Arun Kumar\\
IBM Research\\
\{riddasgu, bganesa1, aswkanna\}@in.ibm.com,
\ reinwald@us.ibm.com
\ kkarun@in.ibm.com
}
\maketitle
\begin{abstract}
Entity Type Classification can be defined as the task of assigning category labels to entity mentions in documents. While neural networks have recently improved the classification of general entity mentions, pattern matching and other systems continue to be used for classifying personal data entities (e.g. classifying an organization as a media company or a government institution for GDPR, and HIPAA compliance). We propose a neural model to expand the class of personal data entities that can be classified at a fine grained level, using the output of existing pattern matching systems as additional contextual features. We introduce new resources, a personal data entities hierarchy with 134 types, and two datasets from the Wikipedia pages of elected representatives and Enron emails. We hope these resource will aid research in the area of personal data discovery, and to that effect, we provide baseline results on these datasets, and compare our method with state of the art models on OntoNotes dataset.
\end{abstract}

\section{Introduction}
In recent years, the increasing emphasis on privacy, and the requirement to comply with regulations have led to the development of data protection systems. These systems have relied on pattern matching and other approaches to detect Personally Identifiable Information (PII), Sensitive Personal Information (SPI), and Protected Health Information (PHI) as per their requirements.

While these have served well so far, the introduction of Global Data Protection Regulation \textbf{(GDPR)} in the European Union (EU) countries has significantly expanded the scope of data protection systems.

Data protection systems involve the extraction of personal data entities (entity recognition), their classification (entity classification) and protection (e.g. encryption, de-identification) based on the sensitiveness of the data. At each of these tasks, data protection systems require very high recall, and reasonable precision. This is because false negatives could lead to loss of private data, while a slight loss in precision because of false positives might be acceptable. Pattern matching and dictionary based systems tend to have higher precision, but need to be continuously updated to achieve good recall. Further, each new type needs to be associated with its own rules and dictionaries, which end up being expensive in terms of money and time and human labour. 

\subsection{Personal Data Entity}
We can define Personal Data Entity (PDE) as any information about a person. Such information can be present in both the public domain as well as in personal data. 
\begin{verbatim}
\end{verbatim}
\begin{itshape}
    Roby was born in Montgomery, Alabama and attended New York University, where she received a bachelor of music degree.
\end{itshape}
\begin{verbatim}
\end{verbatim}
The above sentence is from the publicly available Wikipedia page of an elected official. This sentence by itself cannot be considered as personal data. But it contains Personal Data Entities (PDEs), i.e. entities which are mentioned in a personal context. A news article may also contain such mentions about an elected official.

On the other hand, data in the private domain like emails, chat conversations, medical patient notes, transcripts of voice conversations, employee records can all be considered personal data. For the purpose of our discussion, a mention of a popular person (e.g. actor Matt Damon) in a private conversation should still be considered a PDE mention.

\begin{verbatim}
Montgomery - LOCATION
Alabama - LOCATION
New York University - ORGANIZATION
\end{verbatim}

In the above examples, LOCATION and ORGANIZATION are the labels assigned by the Stanford Named Entity Recognizer (NER). These labels can be considered as \textbf{coarse types} of these entities. Conditional Random Fields (CRF) can be trained to assign a limited number of such labels. However, NER systems can also provide \textbf{fine types} like below.

\begin{verbatim}
ROBY - PERSON
Montgomery - CITY
Alabama - STATE_OR_PROVINCE
\end{verbatim}

These fine types are typically obtained by pattern matching with regular expressions, looking up dictionaries of people names and geographical data, and rule based systems.

The examples shown above happen to be PDEs. However some of the other labels from NERs like NUMBER, ORDINAL, and PERCENT cannot be considered PDE without knowing the context. In fact, even instances of coarse grained labels such as ORGANIZATION cannot be considered to be personal without observing the context in which the entity was mentioned.

In recent years, a number of Neural Fine Grained Entity Classification (NFGEC) models have been proposed, which assign fine grained labels to entities based on context. For example, \textit{New York University} could be typed as \slash{org}\slash{education}.

However the focus of such systems has not been on PDEs. They do not treat the problem of identifying PDEs any different from other entities. For the purpose of GDPR and other regulations, it might be desirable to assign the label \slash{bio}\slash{education}\slash{alma\_mater} to \textit{New York University} and \slash{bio}\slash{education}\slash{edu\_degree} to \textit{bachelor of music}. In contrast to fine grained entity typing systems, standard coarse grained NER systems would have assigned the label \slash{title} to \textit{bachelor of music}, and \slash{organization} to \textit{New York University}.

In this work, we only discuss classifying of PDEs in unstructured data. Personal data entities also occur in structured data, as well as multi-modal data which are beyond the scope of this work. We also do not discuss genome and related biometric data, and leave them for future work. Towards that goal, we can summarize our contributions in this work as follows:
\begin{itemize}
    \item We propose a set of 134 Personal Data Entity Types (PDET), which are fine-grained entity types related to personal data
    \item We introduce 2 new datasets annotated with fine-grained PDETs, which can be used to evaluate PDE typing systems
    \item We propose an approach to improve state of the art models for fine-grained entity classification, by using existing NER systems (hereafter called as Personal Data Annotators) as side information
\end{itemize}

The rest of the paper is organized as follows. We discuss related work, then describe the personal data entity types (PDET) we have created, and explain how annotated two datasets with personal data entities (PDEs). We then discuss our improvements to a state of the art neural model~\cite{shimaoka2017neural} by adding the output of Personal Data Annotators as additional contextual features. Later we briefly explain a PDE Classification pipeline which includes the personal data annotators, the neural model and a post processing step. We then point to future work and conclude with a summary of our findings.

\section{Related Work}
\subsection{Entity Classification}
Entity classification is a well known research problem in Natural Language Processing (NLP).~\cite{ling2012fine} proposed the FIGER system for fine grained entity recognition. In recent years,~\cite{yogatama2015embedding},~\cite{shimaoka2017neural},~\cite{choi2018ultra} have proposed different neural models for context dependent fine grained entity classification.~\cite{abhishek2017fine}~\cite{xu2018neural} proposed improvements to such models using better loss functions.

\cite{dernoncourt2017identification} proposed a RNN model for the de-identification of Protection Health Information. This model has very high F1 on the de-identification task. However, the number of PDE types that can be classified using this approach is limited, as structured prediction and sequence labelling models based on RNNs and CRFs have difficulty scaling up to a large number of classes.

\subsection{Datasets}
\cite{ling2012fine} introduced the  Wiki dataset that consists of 1.5M
sentences  sampled  from  Wikipedia  articles. OntoNotes dataset by~\cite{weischedel2013ontonotes} consists  of 13,109  news  documents  where  77  test  documents are  manually  annotated~\cite{gillick2014context}. BBN dataset by~\cite{weischedel2005bbn} consists of 2,311  Wall  Street Journal  articles  which  are  manually  annotated  using  93  types.~\cite{murty2017finer} have proposed a much larger label set based on Freebase. 

\subsection{Data Loss Prevention}
Entity Classification on Personal Data is much sought after in Big Data and Cloud services. Data Loss Prevention (DLP) systems have used rule based / pattern matching methods to identify personal data.~\cite{wootton2011methods} describe a rule based approach to categorize data in the cloud for DLP. Amazon Macie\footnote{https://aws.amazon.com/macie/}, Google DLP\footnote{https://cloud.google.com/dlp/}, IBM Security Guardium\footnote{https://www.ibm.com/in-en/security/data-security/guardium}, and Microsoft Azure Information Protection\footnote{https://azure.microsoft.com/en-in/services/information-protection/} system are some of the examples of DLP systems.

\section{Personal Data Entity Types (PDET)}
\begin{figure*}
    \includegraphics[width=\textwidth]{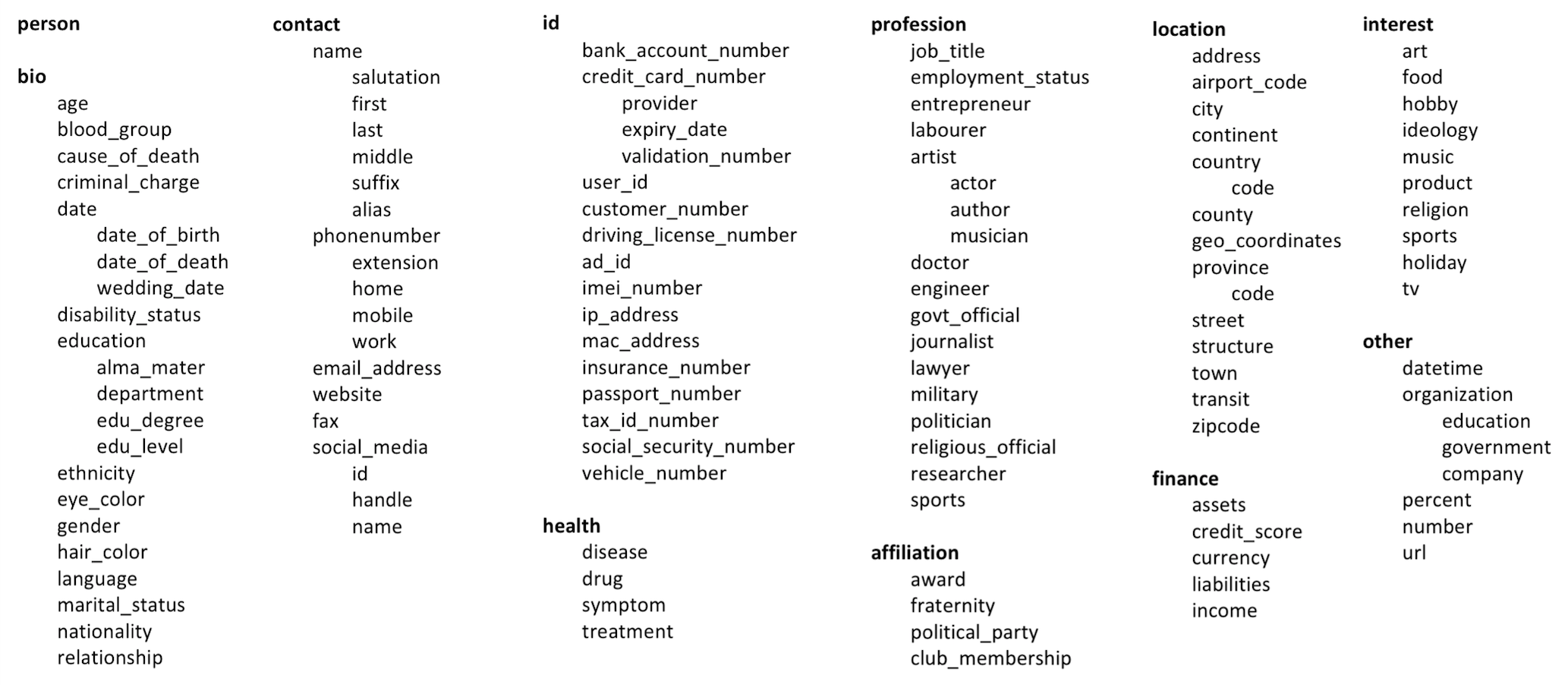}
    \caption{Personal Data Entity Types (PDET)}
    \label{fig:pde_hierarchy}
\end{figure*}
\cite{ling2012fine} proposed the FIGER entity type hierarchy with 112 types.~\cite{gillick2014context} proposed the Google Fine Type (GFT) hierarchy and annotated 12,017 entity mentions with a total of 89 types from their label set. These two hierarchies are general purpose labels covering a wide variety of domains. 
Considering the requirements of GDPR compliance, we propose a larger set of Personal Data Entity Types with 134 entity types as shown in Figure~\ref{fig:pde_hierarchy}.

In order to come up with this hierarchy, we started with the taxonomies proposed by various organizations for GDPR compliance. However such taxonomies include multi-modal data and have substantially more labels than FIGER and GFT. Training a neural model with very large number of class labels may not provide optimal results. Further, obtaining training data for each of the labels was also a concern. Hence we have incorporated a subset of these GDPR taxonomies in our hierarchy.

On the other hand, entity recognition and de-identification models typically have limited number of entity types. We have incorporated all the PHI entity types except biometric entity types. We then considered several NERs, rule based systems and pattern matching systems, and incorporated PDEs recognized by them. We discuss these systems in more detail in the next section. Finally, we included the labels in FIGER and GFT that are relevant to PDEs.

We created the Personal Data Entity Types hierarchy as a stand alone exercise, before considering datasets where entity mentions could be found for training models on this label set. This approach is similar to designing an ontology for a domain, although that is beyond the scope of this work.

\section{PDE Annotators}
Any system that assigns a label to a span of text can be called an annotator. In our case, these annotators assign an entity type to every entity mention. The annotators we have chosen are Stanford Open NLP, and two enterprise (rule/pattern based) annotation systems, IBM BigInsights NER\footnote{\url{https://www.ibm.com/support/knowledgecenter/en/SSPT3X_3.0.0/com.ibm.swg.im.infosphere.biginsights.text.doc/doc/ana_txtan_extractor-libraries.html}} and IBM InfoSphere Information Server\footnote{\url{https://www.ibm.com/analytics/information-server}}.

Stanford Open NLP provides 23 labels, BigInsights provides 18 labels and InfoSphere provides 164 labels. However InfoSphere annotators were written for annotating structured data and need to be provided the spans for entity mentions when dealing with unstructured data.

We use these personal data annotators in 3 ways:
\begin{itemize}
    \item To annotate the two datasets that we are introducing.
    \item To generate the coarse entity types that are used as additional contextual features to our neural model.
    \item As part of the Personal Data Classification pipeline, where for some of the classes, the output of these PDAs are directly used as entity types. These are types like email address, zip codes, number where rule-based systems provide coarse labels at high precision.
\end{itemize}

While neural networks have recently improved the performance of entity classification on general entity mentions, pattern matching and dictionary based systems continue to be used for identifying personal data entities in the industry.

We believe our proposed approach, consisting of modifications to state-of-the-art neural networks, will work on personal datasets for two reasons.~\cite{yogatama2015embedding} showed that hand-crafted features help, and~\cite{shimaoka2017neural} have shown that performance varies based on training data domain. We have incorporated these observations into our model, by using coarse types from rule-based annotators as side information. 

\section{PDE Datasets}
None of the existing fine-grained entity typing datasets have an emphasis on PDEs. As such, in order to evaluate the performance of our proposed approach and the PDE Classification pipeline, we create two new datasets with a focus on fine-grained PDETs. We plan to make these resources available to the community. In this section, we describe our method to create and annotate these datasets.

\begin{table}[!htb]
    \begin{center}
    \begin{tabular}{llll}
        \textbf{}                       & \textbf{Elected Reps} & \textbf{Enron Emails} \\
        \hline
        \textbf{Documents}               & 1196                 & 918 \\
        \textbf{Sentences}               & 38710                & 30941 \\
        \textbf{Entity mentions}         & 211647               & 249278 \\
        \textbf{Unique entity mentions}  & 45686                & 24771 \\
        \textbf{Unique entity types}     & 91                   & 47
    \end{tabular}
    \caption{Statistics of datasets annotated with PDEs}
    \label{tab:datasets}
    \end{center}
\end{table}

As discussed in the introduction section, Personal Data Entities can occur in both publicly available data like Wikipedia pages, as well as in personal data like email conversations. Hence we have created a dataset each from public and personal data, both annotated with PDEs.

\begin{figure*}[!htb]
    \begin{center}
    \includegraphics[scale=0.7]{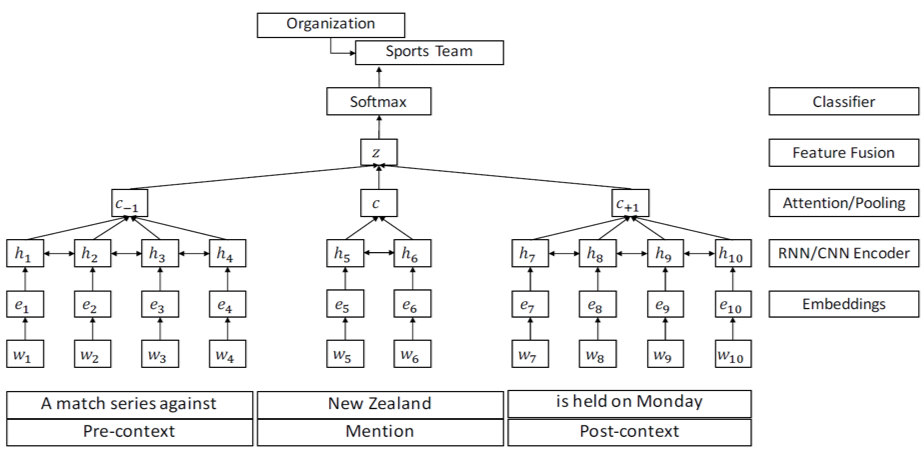}
    \caption{Neural PDE Classification Model Architecture, inspired by~\cite{shimaoka2017neural}}
    \label{fig:nfgec_model}
    \end{center}
\end{figure*}

\subsection{Elected Representatives Dataset}
We have created this dataset from the Wikipedia page of US House of Representatives and the Members of the European Parliament. We obtained the names of 1196 elected representatives from the listings of these legislatures. These listings provide the names of the elected representatives and other details like contact information. However this semi-structured data by itself cannot be used for training a neural model on unstructured data. 

Hence, we first obtained the Wikipedia pages of elected representatives. We then used Stanford OpenNLP to split the text into sentences and tokenize the sentences. We ran the Personal Data Annotators on these sentences, providing the bulk of the annotations that are reported in Table~\ref{tab:datasets}.

We then manually annotated about 300 entity mentions which require fine grained types like \slash{profession}. The semi-structured data obtained from the legislatures had name, date of birth, and other entity mentions. We needed a method to find these entity mentions in the wikipedia text, and assign their column names or manual label as PDEs.

We used the method described in~\cite{chiticariu2010systemt} to identify the span of the above entity mentions in wikipedia pages. This method requires creation of dictionaries each named after the entity type, and populated with entity mentions. 
This approach does not take the context of the entity mentions while assigning labels and hence the data is somewhat noisy. However, labels for name, email address, location, website do not suffer much from the lack of context and hence we went ahead and annotated them.

\subsection{Enron Emails Dataset}
The Enron Corpus\footnote{https://www.cs.cmu.edu/~./enron/} is a database of emails from employees of the Enron Corporation, which was made public for research purposes. We converted 917 Enron emails from the dataset into an appropriate format. We treated the text of the email similar to Wikipedia pages above and annotated PDEs on them. We again used other fields like sender, receiver, timestamp etc on the text of the email to further expand the size of the annotated dataset.

\section{Neural Fine Grained Entity Classification}
Similar to~\cite{shimaoka2017neural} ,~\cite{abhishek2017fine} ,~\cite{choi2018ultra} ,~\cite{murty2017finer} ,~\cite{xu2018neural} ,~\cite{xin2018improving}, we pose fine-grained entity classification as a multi-class, multi-label classification problem, i.e. each sample can belong to multiple labels, which can themselves be multi-class. As the backbone of our architecture, we use the neural network models from~\cite{shimaoka2017neural}, which consists of an encoder for the left and right contexts of the entity mention, another encoder for the entity mention itself, and a logistic regression classifier working on the features from the aforementioned encoders. An illustration of the model is shown in Figure~\ref{fig:nfgec_model}.

The major contribution of~\cite{yogatama2015embedding} was showing the relevance of hand-crafted features for entity classification.~\cite{shimaoka2017neural} further showed that entity classification performance varies significantly based on the input dataset (more than usually expected in other NLP tasks). 

The major drawback of the features used in~\cite{shimaoka2017neural} was the use of custom hand crafted features, tailored for the specific task, which makes generalization and transferability to other datasets and similar tasks difficult. Building on these ideas, we have attempted to augment neural network based models with low level linguistic features which are obtained cheaply to push overall performance. Below, we elaborate on some of the architectural tweaks we attempt on the base model. 

\subsection{Formulation}
Given an entity mention in a sentence, we can rewrite the input as $S = wl_1,...,wl_C, wm_1,...,wm_M, wr_1,...,wr_C$, where $C$ is the windows size for the left and right context, and $M$ is the number of words in the entity mention itself. It can be noted that due to the position of the entity mention, the left or right context can end up being empty, in which case it is replaced with padding. Given this input, the classifier has to predict the labels $t \in \left \{ 0, 1 \right \}^K$. We do this by computing a probability $y_k$ for each possible label $k \in K$. At inference time, the label $k$ with the highest probability, as well as all other labels with $y_k > 0.5$ are predicted.

\subsection{Model Architecture}
Similar to~\cite{shimaoka2017neural}, we use two separate encoders for the entity mention and the left and right contexts. For the entity mention, we resort to using the average of the word embeddings for each word. For the left and right contexts, we employ the three different encoders mentioned in~\cite{shimaoka2017neural}, viz. 
\begin{itemize}
    \item The averaging encoder, which like the mention encoder, and uses the average as the context representation
    \item The RNN encoder, which runs an RNN over the context and takes the final state as the representation of the context
    \item The attentive encoder, which runs a bidirectional RNN over the context, and employs self-attention to obtain scores for each word, which are in turn used to get a weighted sum of the states to use as the representation.
\end{itemize}

Details of the different encoders can be found in~\cite{shimaoka2017neural}, and we omit them here for brevity. The features from the mention encoder, and the left and right context encoders are concatenated, and passed to a logistic regression classifier. If we consider $v_{left}$ to be the representation of the left context, $v_{right}$ to be the representation of the right context, and $v_{entity}$ to be the representation of the entity mention, each being $D$ dimensional then these features are concatenated to form $v = [ v_{left}, v_{right}, v_{entity}$, which is passed to the logistic regression classifier, which in turn computes the function:
\begin{equation}
    y = \frac{1}{1 + exp \left ( -W_y v \right )}
\end{equation}
where $W_y$ is the set of weights that project the features from a $3 \times D$ dimensional feature space to a $K$ dimensional output, where $K$ is the number of labels, and $0 \leq y_k \leq 1 \forall k \in K$. Since the output is a binary vector, we employ a binary cross entropy loss during training. Given the predictions $y$ and the ground truth $t$ for a sample, the loss is defined as:
\begin{equation}
    L(y, t) = \sum_{k=1}^{K} -t_k log(y_k) - (1-t_k)log(1-y_k)
\end{equation}
We employ stochastic mini-batch gradient descent to optimize the above loss function, and the details are specified later in the experimental results section.

\subsection{Embeddings}
\begin{figure}[!htp]
    \begin{center}
    \includegraphics[width=0.7\columnwidth]{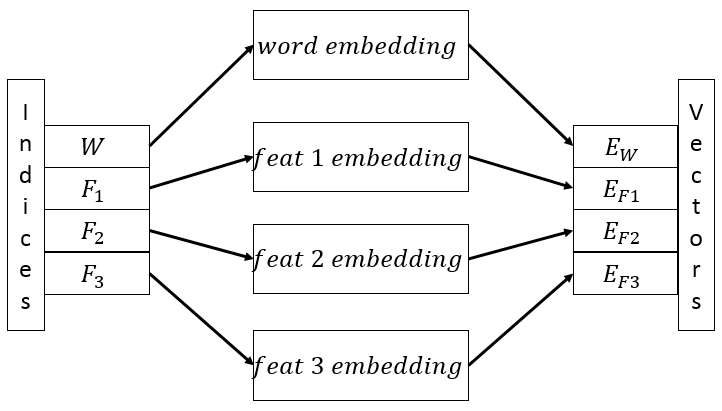}
    \caption{Illustration of how feature embeddings are concatenated with word embeddings}
    \label{fig:embeddings}
    \end{center}
\end{figure}
The input to our model is a sequence of words, represented by their corresponding embeddings by a look up table. Traditionally, pre-trained word embeddings such as GloVe~\cite{pennington2014glove} are used. Earlier work such as~\cite{shimaoka2017neural} have kept the word embeddings frozen during training, but we update them, to account for words that might be present in our datasets but not in the GloVe vocabulary. Our main contribution comes in the form of augmented embeddings, wherein we concatenate embeddings for token level features to the word embedding. Each word can also be represented in a plethora of ways, such as using POS tags, dependency parse tags, NER tags, etc. We peruse a few of these cheaply available annotations, project them to a low dimensional embedding space, and concatenate the said embeddings to the word embedding. For a word $W$, whose word embedding is denoted by $E_W$, with features $F_1, ..., F_N$, whose embeddings are denoted by $E_{F1}, ..., E_{FN}$, the final embedding is given by $[ E_W, E_F1, ..., E_FN ]$. A pipeline of how to construct the embeddings is shown in Figure~\ref{fig:embeddings}.

\section{Experimental Results}
\begin{table}[!htb]
    \begin{center}
    \begin{tabular}{llll}
        \textbf{Dataset}                 & \textbf{\# Test samples} & \textbf{\# Labels} \\
        \hline
        \textbf{OntoNotes}               & 8963                     & 89 \\
        \textbf{Elected Representatives} & 16805                    & 91 \\
        \textbf{Enron Emails}            & 15960                    & 47 
    \end{tabular}
    \caption{Statistics of the datasets used in our experiments.}
    \label{tab:dataset_details}
    \end{center}
\end{table}
\vspace{-5mm}
\subsection{Datasets}
For our experiments, we leverage the widely used OntoNotes dataset~\cite{gillick2014context}, as well as the Elected Representatives and Enron Emails datasets that we curated ourselves. Table~\ref{tab:dataset_details} contains the details of the datasets, including train/test splits sizes, as well as number of fine-grained entities in each dataset.

\begin{table*}[!htb]
    \begin{center}
    \begin{tabular}{cccccc}
    \hline
    \textbf{Encoder} & \textbf{Setting} & \textbf{Accuracy} & \textbf{Macro F1} & \textbf{Micro F1} & \textbf{Gmean} \\
    \hline
    \multirow{3}{*}{\textbf{Avg}} & \textbf{Paper}          & 0.462    & 0.653    & 0.582     & 0.559 \\
                                  & \textbf{Our Baseline}   & 0.481    & 0.678    & 0.617     & 0.586 \\
                                  & \textbf{PDA Features}   & 0.534    & 0.740    & 0.672     & 0.643 \\
                                  \cline{2-6}
    \multirow{3}{*}{\textbf{Rnn}} & \textbf{Paper}          & 0.492    & 0.667    & 0.605     & 0.583 \\
                                  & \textbf{Our Baseline}   & 0.494    & 0.693    & 0.635     & 0.602 \\
                                  & \textbf{PDA Features}   & 0.537    & 0.737    & 0.67      & 0.642 \\
                                  \cline{2-6}
    \multirow{3}{*}{\textbf{Att}} & \textbf{Paper}          & 0.503    & 0.679    & 0.616     & 0.595 \\
                                  & \textbf{Our Baseline}   & 0.493    & 0.677    & 0.612     & 0.589 \\
                                  & \textbf{PDA Features}   & 0.543    & 0.743    & 0.675     & 0.648 \\
    \hline
    \end{tabular}
    \caption{Performance of adding features embeddings to word embeddings for OntoNotes dataset. GMean denotes the geometric mean of accuracy, macro F1 and micro F1 scores. PDA Features refers to POS tags, NERs and annotations from rule-based annotators. Paper refers to the original numbers as reported in~\cite{shimaoka2017neural}.}
    \label{tab:ontonotes_results}
    \end{center}
\end{table*}

\subsection{Hyperparameters}
We used a standard set of hyperparameters for most of our experiments. Optimal values of learning rate and batch size were obtained by evaluating model performance on held out validation splits. In a departure from previous methods such as~\cite{shimaoka2017neural},~\cite{abhishek2017fine}, which use large mini-batches of $~1000$ samples, we use a smaller batch size of 512 samples, after trying out batch sizes of $128, 256, 512, 1024$. We also use an appropriate learning rate of $0.0001-0.0005$, in conjunction with the Adam optimizer~\cite{kingma2014adam}. Following~\cite{shimaoka2017neural}, we use a dropout~\cite{srivastava2014dropout} of $0.5$ as regularizer on the encoders. The context window length is also set to $10$, and padding is used if the context is smaller. $300$-dimensional GloVe vectors were used, and words not found in the GloVe vocabulary were initialized randomly and learnt during training. For the RNN and attentive encoders, the LSTM hidden size was set to $100$. Feature embeddings were set to $16$ dimensions. POS tags and NER features were obtained using Stanford Core NLP, while Type Tags were obtained using the rule based annotation system InfoSphere mentioned earlier. To evaluate performance, we follow~\cite{ling2012fine} and use accuracy or strict F1 score, macro averaged F1 score, and micro averaged F1 score. To compare across different runs, we use the geometric mean of the 3 different metrics. 

\subsection{Influence of Token Level Features}
\begin{table}[!htb]
    \begin{center}
    \begin{tabular}{lccccc}
    \textbf{Features}    & \textbf{Acc.}     & \textbf{Ma-F1}    & \textbf{Mi-F1}    & \textbf{GMean} \\
    \hline
    \textbf{None}        & 0.481             & 0.678             & 0.617             & 0.586 \\
    \textbf{Pos}         & 0.503             & 0.706             & 0.633             & 0.608 \\
    \textbf{Ner}         & 0.504             & 0.720             & 0.649             & 0.618 \\
    \textbf{Typ}         & 0.517             & 0.736             & 0.667             & 0.633 \\
    \textbf{Pos+Ner}     & 0.512             & 0.727             & 0.653             & 0.624 \\
    \textbf{Pos+Typ}     & 0.507             & 0.732             & 0.658             & 0.625 \\
    \textbf{Ner+Typ}     & 0.514             & 0.730             & 0.661             & 0.628 \\
    \textbf{Pos+Ner+Typ} & 0.534             & 0.740             & 0.672             & 0.643 \\
    \end{tabular}
    \caption{Performance of adding features embeddings to word embeddings for OntoNotes dataset. GMean denotes the geometric mean of accuracy, macro F1 and micro F1 scores. Pos refers to POS tags, Ner refers to rudimentary named entities, while Typ refers to annotations from rule-based annotators.}
    \label{tab:ontonotes_ablation}
    \end{center}
\end{table}
Table~\ref{tab:ontonotes_results} shows how our proposed architectural changes at the embedding level improve performance across all metrics when compared to the base model with plain word embeddings from~\cite{shimaoka2017neural}. The first row, titled $Paper$ for each encoder, denotes the original results as reported by~\cite{shimaoka2017neural}. Results from our re-implementation, which updates word embeddings during training and uses a smaller batch size of $512$, are highlighted in the rows titled $Our Baseline$. The final row in each encoder section, titled $PDA Features$ shows the effect of concatenating token level features using Personal Data Annotators. As is evident, these features always improve the performance irrespective of the type of encoder.

Table~\ref{tab:ontonotes_ablation} showcases the performance of concatenating feature embeddings to the pre-trained word embeddings. Since we have 3 different types of features, viz. POS/NER/TYP, we perform a complete ablation analysis of the influence of each feature. We only display results from the averaging encoder for brevity, although similar trends were observed across all encoders. In the table, Pos refers to POS tags, Ner refers to coarse named entities, while Typ refers to annotations from rule-based annotators. The first row, with $None$ features, is the baseline, while the remaining rows highlight the efficacy of adding POS tags, NERs and Type tags to the pre-trained word embeddings. As can be seen, NER and Type tags have the highest influence on fine-grained entity classification. These results support our hypothesis that token level features, specially coarse grained NERs and Type tags from rule based systems, aid fine grained typing of entity mentions with context. 

\subsection{Performance on PDE datasets}
\begin{table*}[!htb]
    \begin{center}
    \begin{tabular}{ccccccc}
    \hline
    \textbf{Dataset} & \textbf{Encoder} & \textbf{Setting} & \textbf{Accuracy} & \textbf{Macro F1} & \textbf{Micro F1} & \textbf{Gmean} \\
    \hline
    \multirow{6}{*}{\textbf{\shortstack[c]{Enron\\Emails}}} 
                           & \multirow{2}{*}{\textbf{Avg}}  & \textbf{Baseline} & 0.957 & 0.981 & 0.979 & 0.972 \\
                           &                                & \textbf{Features} & 0.986 & 0.995 & 0.994 & 0.992 \\
                           \cline{2-7}
                           & \multirow{2}{*}{\textbf{Rnn}}  & \textbf{Baseline} & 0.960 & 0.981 & 0.979 & 0.973 \\
                           &                                & \textbf{Features} & 0.985 & 0.995 & 0.993 & 0.991 \\
                           \cline{2-7}
                           & \multirow{2}{*}{\textbf{Att}}  & \textbf{Baseline} & 0.960 & 0.981 & 0.979 & 0.973 \\
                           &                                & \textbf{Features} & 0.987 & 0.995 & 0.994 & 0.992 \\
    \hline
    \multirow{6}{*}{\textbf{\shortstack[c]{Elected\\Representatives}}}
                           & \multirow{2}{*}{\textbf{Avg}}  & \textbf{Baseline} & 0.903 & 0.959 & 0.955 & 0.939 \\
                           &                                & \textbf{Features} & 0.964 & 0.989 & 0.985 & 0.979 \\
                           \cline{2-7}
                           & \multirow{2}{*}{\textbf{Rnn}}  & \textbf{Baseline} & 0.900 & 0.958 & 0.953 & 0.936 \\
                           &                                & \textbf{Features} & 0.963 & 0.989 & 0.985 & 0.979 \\
                           \cline{2-7}
                           & \multirow{2}{*}{\textbf{Att}}  & \textbf{Baseline} & 0.899 & 0.958 & 0.953 & 0.936 \\
                           &                                & \textbf{Features} & 0.963 & 0.989 & 0.985 & 0.979 \\
    \hline
    \end{tabular}
    \caption{Performance on Elected Representatives and Enron Emails datasets. GMean denotes the geometric mean of accuracy, macro F1 and micro F1 scores. $Baseline$ refers to the same setup as in~\cite{shimaoka2017neural}, without any features, whereas $Features$ refers to having POS tags, NERs and annotations from rule-based annotators.}
    \label{tab:enron_elected_features}
    \end{center}
\end{table*}
The results on Elected Representatives and Enron Emails dataset, which can be seen in table~\ref{tab:enron_elected_features}, clearly show the same trend, i.e. adding token level features improve performance across the board, for all metrics, as well as for any choice of encoder. The important thing to note is that these token level features can be obtained cheaply, using off-the-shelf NLP tools to deliver linguistic features such as POS tags, or using existing rule based systems to deliver task or domain specific type tags. This is in contrast to previous work such as~\cite{ling2012fine},~\cite{yogatama2015embedding} and others, who resort to carefully hand crafted features.

\subsection{Class Wise Performance}
\begin{table}[]
    \begin{center}
    \begin{tabular}{lcc}
    \textbf{Label}                      & \textbf{Baseline F1} & \textbf{Features F1} \\
    \hline
    \textbf{/location/city}             & 0.24                 & 0.52 \\
    \textbf{/org/company/news}          & 0.0                  & 0.49 \\
    \textbf{/person/political\_figure}  & 0.08                 & 0.14 \\
    \textbf{/person/title}              & 0.29                 & 0.44 \\
    \textbf{/person/artist}             & 0.07                 & 0.09                
    \end{tabular}
    \caption{F1 scores for some labels from OntoNotes for both the baseline model, without features, as well as the model with PDA features.}
    \label{tab:class_wise_metrics}
    \end{center}
\end{table}
In table~\ref{tab:class_wise_metrics}, we show the class-wise F1 scores for some select classes in the OntoNotes dataset. As can be seen, performance clearly improves with the addition of token-level Personal Data Annotators features. Similar trends can be observed for labels in the other PDE datasets as well. Note that the classes highlighted are all fine-grained classes, which highlights the efficacy of the proposed PDA features for the task of fine-grained personal data entity classification.

\section{PDE Classification Pipeline}
\begin{figure}[ht]
    \includegraphics[width=\columnwidth]{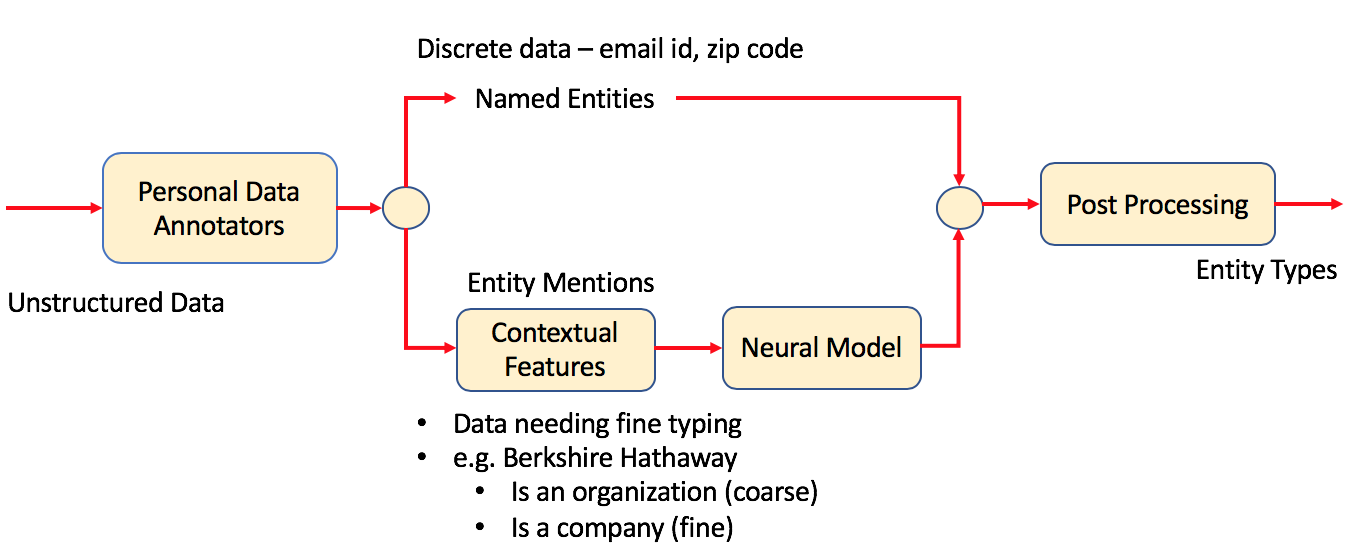}
    \caption{Personal Data Entities classification pipeline}
    \label{fig:pipeline}
\end{figure}

We have implemented a pipeline for Personal Data Entity Classification as shown in Figure~\ref{fig:pipeline}. This pipeline consists of existing personal data annotators, the neural fine grained entity classification model described in the previous section, and a rule-based post processing step to combine the output of rule-based annotators and the neural model.

The input to our pipeline are text sentences. We use existing entity recognizers to find mentions. The output is a list of fine grained entity types for each of the mentions. We have a rule based system to post process the results from both the Personal Data Annotators and the neural model.

\section{Future Work}
In PDE classification, there are still a number of open problems. We mention some of them here. Using \textbf{co-reference resolution} or other approaches to determine, for example who is the doctor and the patient in a medical patient note could be a useful addition to this work. A downstream anonymization solution can choose to redact the patient name, while leaving the doctor's name intact. In many applications, the ability to peruse the document for analytics after anonmyization is considered important.

Another potential improvement is \textbf{generalizing the model} to work on any domain, as long as we have some rule-based coarse level annotators, and training data at fine grained level. For example, patient notes and other data in health care domain can be annotated with NLM Scrubber tool from~\cite{kayaalp2015easy}.

In this work, we have focused only on unstructured data. This work can also be extended to \textbf{PDE classification on structured data}. This can be approached in two ways. Deep Learning for Tabular data has recently begun to gain traction and can be attempted for the PDE classification task. Another approach could be to generate context from meta-data and other columns similar to unstructured data.

\section{Conclusion}
We introduced Personal Data Entities (PDE) as a separate set of entities that need be classified differently than general fine grained entity classification. We introduced a hierarchy of 134 Personal Data Entity Types (PDET), and described two datasets annotated with PDEs. We then proposed an approach to use existing rule-based annotators, to generate additional context features for a state of the art neural. Our experiment results show a substantial increase in accuracy, micro and macro F1 over the baseline model.


\bibliographystyle{aaai}
\bibliography{aaai_bib_file}
\end{document}